\documentclass{article}

\usepackage[final]{neurips_distshift_2021}

\usepackage[utf8]{inputenc} % allow utf-8 input
\usepackage[T1]{fontenc}    % use 8-bit T1 fonts
\usepackage{hyperref}       % hyperlinks
\usepackage{url}            % simple URL typesetting
\usepackage{booktabs}       % professional-quality tables
\usepackage{amsfonts}       % blackboard math symbols
\usepackage{nicefrac}       % compact symbols for 1/2, etc.
\usepackage{microtype}      % microtypography
\usepackage{xcolor}         % colors

\usepackage{enumerate}
\usepackage{graphicx}
\usepackage[ruled,vlined]{algorithm2e}
\usepackage{bm}
\newcommand{\fedavg}{\textsc{FedAvg}\xspace}
\newcommand{\clientopt}{\textsc{ClientOpt}\xspace}
\newcommand{\lr}{\eta}
\newcommand{\serveropt}{\textsc{ServerOpt}\xspace}
\newcommand{\slr}{\lr_{s}}
\newcommand{\vx}{\bm{x}}
\newcommand{\sgrad}{g}
\newcommand{\activeBatch}{\mathcal{B}}
\newcommand{\activeClients}{\mathcal{S}}
\newcommand{\localStep}{\tau}
\newcommand{\localChange}{\Delta}
\newcommand{\centralopt}{\textsc{CentralOpt}\xspace}
\newcommand{\clr}{\lr_{c}}
\newcommand{\mergeopt}{\textsc{MergeOpt}\xspace}
\newcommand{\mlr}{\lr_{m}}
\newcommand{\mergedChange}{\mathbf{\Delta}}

\title{Jointly Learning from Decentralized (Federated) and Centralized Data to Mitigate Distribution Shift}

\author{%
  Sean Augenstein \\
  Google Inc.\\
  \texttt{saugenst@google.com} \\
  \And
  Andrew Hard \\
  Google Inc.\\
  \texttt{harda@google.com} \\
  \AND
  Kurt Partridge \\
  Google Inc.\\
  \texttt{kep@google.com} \\
  \And
  Rajiv Mathews \\
  Google Inc.\\
  \texttt{mathews@google.com} \\  
}

\begin{document}

\maketitle

\begin{abstract}
With privacy as a motivation, Federated Learning (FL) is an increasingly used paradigm where learning takes place collectively on edge devices, each with a cache of user-generated training examples that remain resident on the local device. These on-device training examples are gathered in situ during the course of users' interactions with their devices, and thus are highly reflective of at least part of the inference data distribution. Yet a distribution shift may still exist; the on-device training examples may lack for some data inputs expected to be encountered at inference time. This paper proposes a way to mitigate this shift: selective usage of datacenter data, mixed in with FL. By mixing decentralized (federated) and centralized (datacenter) data, we can form an effective training data distribution that better matches the inference data distribution, resulting in more useful models while still meeting the private training data access constraints imposed by FL.
\end{abstract}

\section{Introduction and Background}
\label{intro}

Federated learning (FL)~\citep{mcmahan2017fedavg} is a machine learning setting where multiple `clients' (typically, edge computing devices like mobile phones) collaborate in solving a machine learning problem, under the coordination of a central server. Each client caches raw training data locally, and the data are never exchanged or transferred. Instead, focused updates intended for immediate aggregation are used to achieve the learning objective~\citep{kairouz2019advances}. Providing strong privacy protection is a major motivation for FL. Storing data locally rather than replicating it in the cloud decreases the attack surface of the system, and the use of focused ephemeral updates and early aggregation follow the principle of data minimization~\citep{whitehouse13privacy}. Stronger privacy properties are possible when FL is combined with technologies such as differential privacy (DP) and secure multiparty computation (SMPC)~\citep{wang2021fedopt}.

FL also typically delivers utility improvements such as increased task accuracy, because the training examples gathered in situ by edge devices are reflective of actual inference serving requests. For example, a mobile keyboard next word prediction model can be trained from actual SMS messages, yielding higher accuracy than a similar model trained instead on a proxy document corpus. Because of these benefits, FL has been adopted and used to train production models for a variety of mobile phone applications, such as keyboard suggestions and audio keyword spotting~\citep{hard2018nwp, ramaswamy2019emoji, apple2019wwdc, hard2020keyword, ramaswamy2020dpnwp, hartmann2021smartselect}.

While FL reduces train vs. inference distribution skew, it doesn't remove it completely. The FL training examples may still not fully represent the inference distribution, due to situations such as:
\begin{itemize}
  \item Label-biased example retention skewing the cached dataset away from the inference distribution. The nature of users' interactions with their devices often deters the caching of negative examples. For example, a mobile phone's photo cache reflects photographic subjects a user is interested in. It generally does not contain photos of subjects a user is \emph{not} interested in (such photos were either not taken, or were taken and deleted). If desiring e.g. a model that predicts interest in a camera scene, the lack of negative examples (i.e., the label imbalance) would be a significant impediment to training a binary `interest' classifier. %For example, a mobile application that triggers via audio keyword is built to capture positive utterances (e.g., `Hey Alexa', `Hey Google', `Hey Siri'), and avoid triggering on or retaining negative examples like non-keyword utterances or random ambient noise. The resulting label imbalance is an impediment.
  \item A model that trains on one class of edge devices but is used for inference on other classes of devices as well. E.g., an audio model may train on mobile phones but also be deployed for inference on smart speakers (with different microphone characteristics), or train on high-end phones but also be deployed for inference on low-end phones (with different user characteristics).
  \item Safety critical applications of AI. Risky scenarios requiring extremely accurate inference behavior are (hopefully) infrequently encountered by users in the course of typical day-to-day device interaction (and thus lacking in the federated training examples). E.g., the intelligent driving assistance prompts to the driver of an automobile skidding off a road in heavy rain.
\end{itemize}

How can the benefits of FL (improved privacy and utility) be achieved while overcoming the remaining train vs. inference distribution skew? This paper proposes a solution: mixing data from an additional datacenter dataset into a FL training process, to afford a `composite' set of training data that better matches the inference distribution. The process for mixing is not trivial, given the stringent privacy requirements of FL. In traditional datacenter-based learning, a ML engineer can simply concatenate two (or more) sets of training data into one, randomly shuffle, and then draw batches of mixed examples. But this level of fine-grained interaction with training examples is precluded in FL, where the on-device examples (containing user private information that reveals identity) stay on device and can never be handled individually by the ML engineer~\citep{augenstein2019generative}.

This paper discusses strategies for training models which meet the private data access constraints imposed by FL while also leveraging additional data available at the server, to achieve models that perform better on the inference distribution. Section~\ref{sec:strategies} presents several such mixing strategies, Section~\ref{sec:experiment} demonstrates their application in an experiment involving the CelebA dataset, Section~\ref{sec:comparisons} compares their performance at several modeling objectives, and Section~\ref{sec:applications} discusses some perspectives and logical extensions in mixing decentralized and centralized data.

\section{Strategies to Learn on Decentralized and Centralized Data}
\label{sec:strategies}

We now consider three strategies that perform FL while mixing in additional data from a centralized dataset. We use $\fedavg$~\citep{mcmahan2017fedavg} (Algorithm~\ref{algo:generalized_fedavg}), a ubiquitous FL algorithm, as the learning approach of reference. Each of the strategies below is a modification of $\fedavg$. The algorithms for the strategies are presented in Appendix~\ref{app:algos}. % We demonstrate the application of the strategies to an example problem using the CelebA dataset in the next section.

\paragraph{Synchronous Parallel Training} In synchronous parallel training, we start with a global model, and then separately in parallel perform a round of \fedavg (with the decentralized data) and steps of training (with the centralized data). This yields two updated versions of the model, one via FL and one via central training. After every round, we take these two versions of the model, merge them together (e.g., average the weights) to form a new global model, and repeat. See Algorithm~\ref{algo:parallel_training} for details.

\paragraph{Example Transfer} In example transfer, we perform standard \fedavg with one additional aspect. For each client participating in the FL round, we sample some training examples from the centralized data and send these to the clients (along with the latest version of the global model). These datacenter examples are merged and shuffled with the local client examples, and then batches of `mixed' data are drawn when performing the client training portion of standard \fedavg. See Algorithm~\ref{algo:ex_transfer} for details. In this way, the client optimization accounts for both decentralized data and centralized data as it proceeds towards a minima.

\paragraph{Gradient Transfer} In gradient transfer, at the start of a round, we first take the latest version of the global model and calculate a gradient w.r.t.~a batch of datacenter training examples. This gradient is then sent to the participating clients (along with the latest version of the global model). During each step of client training, we calculate a gradient w.r.t.~batches of client data examples, sum this gradient with the `augmenting' datacenter gradient, and then use the summed gradient when performing client optimization. See Algorithm~\ref{algo:grad_transfer} for details. In this way, the client optimization is `biased' in a direction such that it is minimizing of both decentralized data and centralized data.

\paragraph{Related Work} In terms of algorithmic approach, there are parallels between some of the above strategies and algorithms for addressing inter-client data heterogeneity in FL. For example, the example transfer strategy is reminiscent of a data-sharing strategy proposed in \citet{zhao2018noniid}. That work aimed to address the problem of optimization on non-IID client datasets, whereas we apply the concept to mixing between the federated clients' overall (decentralized) distribution and a separate centralized distribution. Similarly, the gradient transfer strategy is inspired by the SCAFFOLD~\citep{karimireddy2019scaffold} and Mime~\citep{karimireddy2020mime} algorithms. These algorithms calculate a gradient term that is reflective of the client population as a whole, and transmit it to the clients each round to reduce the variance in the respective client updates. By contrast, the gradient term here is derived from the centralized data, and not the decentralized federated data.

\paragraph{\textit{Negative Strategy: Transfer Learning}} One strategy attempted and ultimately abandoned was transfer learning, a.k.a. fine-tuning. In this approach, a model is pre-trained on centralized data and then fine-tuned on decentralized data via FL. This sequential approach results in `catastrophic forgetting'~\citep{mccloskey1989catforget, ratcliff1990catforget, french1999catforget}; prediction accuracy on the centralized data distribution is lost as the model learns to fit the decentralized data instead. We seek strategies yielding good inference performance against \emph{all} data distributions trained on.

% TODO(saugenst): Reference Aritra Mitra's paper?

\section{Experiment}
\label{sec:experiment}

\paragraph{Scenario} Consider a mobile phone camera portrait application, with a feature desired that would identify if human subjects in photos are smiling (e.g., to suggest to users if a portrait photo should be kept or needs retaking). In the FL paradigm, models are trained on actual user images on device (under stringent user privacy protections, e.g. the images never leave the device). However, as the application usage is one that generally results in only positive (i.e., smiling) example portraits being retained by users, training a binary `smiling'-vs.-`unsmiling' classifier from only the edge device data will be challenging due to label imbalance. We can leverage the fact that many large public corpora of photos of human faces exist, containing many images of unsmiling subjects. We examine mixing this public data into our FL training process, using the strategies presented in Section~\ref{sec:strategies}.

\paragraph{Datasets} We use a version of the CelebA dataset~\citep{liu2015faceattributes}, made into a federated dataset\footnotemark ~by partitioning images up by portrait subject~\citep{caldas2018leaf}. I.e., we treat each celebrity as having a mobile phone with a cache of photos of themselves, which participates as a client in FL. The CelebA data contains a boolean attribute indicating whether a portrait subject is smiling or not. We filter on this attribute, to form two resulting groups of data: a federated dataset partitioned by celebrity, containing only smiling images, and a centralized dataset containing all unsmiling images together (removing partitions and putting all celebrities together in a single set). The former represents the (non-IID) on-device private user data, and the latter is a publicly available corpus of unsmiling faces.

\footnotetext{The CelebA federated dataset is available via open source FL software~\citep{tff2021celeba}.}

\paragraph{Setup} We use the federated and centralized datasets to learn a smiling classifier, under five scenarios: 
\begin{enumerate}[i]
  \item No Mixing, pure \fedavg (Algorithm~\ref{algo:generalized_fedavg}) (no access to any unsmiling examples)
  \item Mixing via Parallel Training (Algorithm~\ref{algo:parallel_training})
  \item Mixing via Example Transfer (Algorithm~\ref{algo:ex_transfer})
  \item Mixing via Gradient Transfer (Algorithm~\ref{algo:grad_transfer})
  \item No Mixing, pure \fedavg, but with all data federated, both smiling and unsmiling examples (i.e., as if celebrities also retained unsmiling portraits on their mobile devices) 
\end{enumerate}

Scenario i is a (baseline) case of training a binary classifier with no access to negative examples. Scenario v provides `oracle' performance; it violates the fundamental assumption that negative examples are not present on user devices, but shows what pure FL performance would look like were this not the case. Scenarios ii-iv each cover a mixing strategy described in Section~\ref{sec:strategies}.

Of the 9343 total clients in federated CelebA, we train on a population of 8408 clients (approximately $90\%$ of the total), and then evaluate on a separate population of the 935 remaining clients. We train for 5000 federated rounds, randomly selecting 100 training clients per round.

Our evaluation data consists of both smiling and unsmiling faces, and is meant to stand in for the inference distribution (where accurate classification of both smiling and unsmiling inputs is necessary). Note that CelebA contains smiling and unsmiling faces in nearly equal amounts, so a high evaluation accuracy cannot come at the expense of one particular label being poorly classified.

\begin{figure}
\centering
\includegraphics[width=1.0\columnwidth]{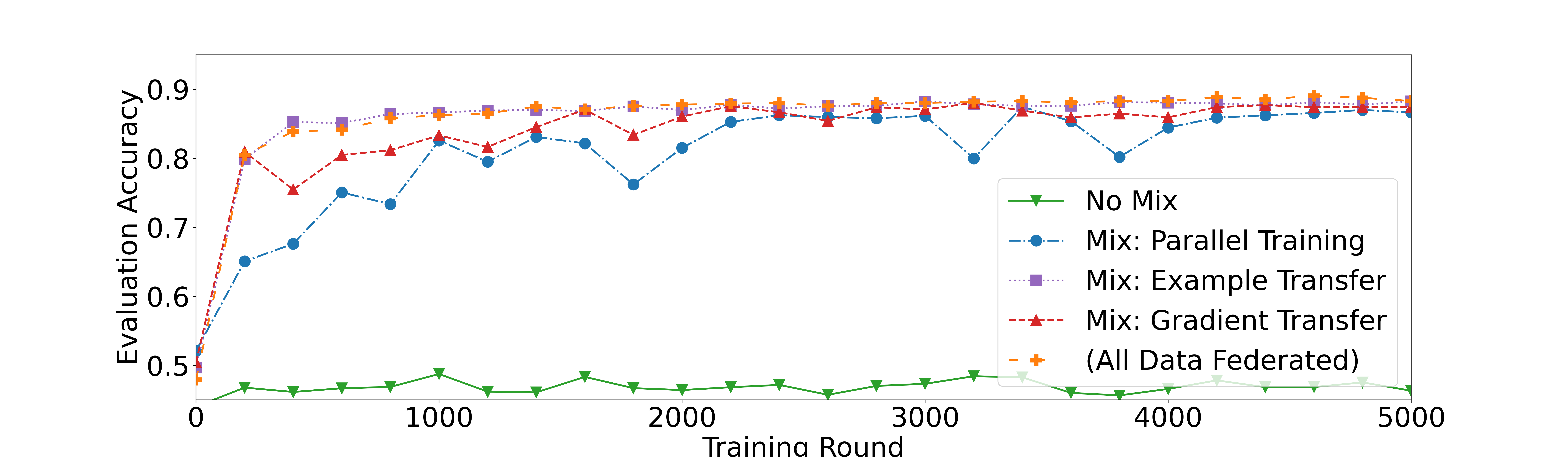}
\caption{CelebA smiling classifier, evaluation accuracy vs. federated training round.}
\label{fig:celeba_accuracy}
\end{figure}

\paragraph{Results} Figure~\ref{fig:celeba_accuracy} shows the classifier evaluation accuracy resulting from each strategy. The three mixing strategies all result in a smiling classifier that is far more accurate than the `no mix' scenario.

% XMgr link for experiment that gathered this data: http://xids/30119976
% Colab script for generating plot: http://colab/drive/1hKtHQRRGkgk2v2G7HF-zOp_OMsgvZxtq?resourcekey=0-KHdDVFD4HJT469YgKnEokw#scrollTo=svnFNkrfG9ei

\section{Comparisons}
\label{sec:comparisons}

There are tradeoffs between the strategies, based on accuracy, transfer payload size, and privacy. 

\paragraph{Accuracy} Differences exist in the resulting accuracy of a classifier trained by the various mixing strategies. This is shown for example in Figure~\ref{fig:celeba_accuracy}, for the smile classifier described above. Example transfer could be expected to perform best, effectively equivalent to a scenario where there were no label imbalance issues and one could just perform pure FL with smiling \emph{and} unsmiling faces. Gradient transfer could be expected to sometimes perform worse than example transfer, for the reason that the gradient w.r.t.~centralized data is calculated once and then frozen over the course of the clients' optimization steps. It may grow `stale' after several steps, as compared to example transfer where the gradient w.r.t.~centralized data is calculated fresh with every step of optimization (because centralized data examples exist on device). In general, further analysis is necessary to understand the relative convergence behavior of these mixing strategies. Also, note that parallel training has a larger hyperparameter space to explore in understanding its optimal converge settings.
 
\paragraph{Transfer payload size} Example transfer and gradient transfer both require transmitting additional information (along with the latest model checkpoint) from server to clients at the start of each federated round. Naively, example transfer adds an amount to the payload of the order of example size times number of examples, and gradient transfer doubles the transfer payload (as the gradient is the same size as the model). Studying compression strategies would be a useful future research exploration, in particular for gradient transfer, as it seems likely that not all gradient terms are equally necessary. Parallel training incurs no additional transfer cost above pure FL. %, so while parallel training achieves the lowest accuracy of the mixing strategies it may still be useful in situations where transfer time constraints dominate. 

\paragraph{Privacy} None of the three mixing strategies send additional information back from the clients to the server, so there are no added concerns (above those in standard \fedavg) with respect to protecting user private data on the clients. The same tools for protecting user data during FL, like user-level DP~\citep{mcmahan2018dplm}, can be brought to bear. Both example transfer and gradient transfer broadcast information about the centralized training data from the server to the clients, but in many scenarios the centralized data is not privacy sensitive (e.g., it is a public corpus). Were the centralized training data to also be privacy sensitive, then that imposes some constraints on what is possible. Example transfer would be precluded, unless heavily modified to an intricate approach that utilizes DP generative models to synthesize privacy-preserving examples~\citep{augenstein2019generative}. Gradient transfer is less directly revealing of the raw centralized examples then example transfer, but recent work has shown that reconstructing sensitive information from gradients alone is possible~\citep{dang2021speakeridattacks}. The same work demonstrates methods for defending against such an attack, which could be adopted if applying gradient transfer to mixing problems involving sensitive centralized data.

\section{Discussion and Further Directions}
\label{sec:applications}

This paper has described and compared basic mixing strategies that result in models that perform better at inference time. We now consider some perspectives on mixing centralized and decentralized data, which inform extensions to the basic strategies presented here.

One perspective on mixing: it is a merging of `wisdom of the crowd' (from decentralized, user private data) with `wisdom of laboratory experts' (from an organized, purposeful data collection exercise). The `crowd' training distribution conveys breadth, while the `expert' training distribution conveys depth at critical regions of the support of the inference data distribution. For example, for a safety critical AI application like a self-driving car, a useful model should meld the massive amount of data gathered by everyday drivers under common conditions with the purposefully collected data gathered by automotive safety organizations using professional drivers on closed tracks under adverse conditions (or high-fidelity simulations thereof). Also, it should do so without compromising the privacy of everyday drivers.

In this vein, we can look at the mixing described in Section~\ref{sec:strategies} as simplistic. It assumes that a simple \emph{union} of two datasets (the decentralized training examples on edge devices and the centralized examples in a datacenter) can usefully match an inference distribution. But what if we need to be more intricate and selective in our use of e.g. centralized training data? 

For example, our centralized corpus might be enormous, with most examples not germane to our immediate problem. Rather than use the entire corpus, we'd like to filter and use only the centralized data examples that are most relevant. E.g., we have access to a public corpus of photos, but wish to filter to just portraits. Alternatively, if lacking any centralized corpus, we might undertake a targeted data collection exercise. E.g. (along the lines of the scenario in Section~\ref{sec:experiment}), all the federated data are smiling faces and we realize we need to organize the paid collection of portraits of unsmiling faces.

To undertake such filtering/targeting, we could use information from the federated private data examples (properly anonymized) as input. We wish to know enough about the federated training data distribution to understand what public data would be germane (e.g., portraits of human faces) but complementary (e.g., portraits with unsmiling subjects) to the decentralized data. Two tools for doing this, which could be combined with mixed learning, are federated analytics~\citep{ramage2020fedanalytics} and federated private generative models~\citep{augenstein2019generative}.

It is also interesting to consider FL applications where privacy is not the dominant concern. Imagine the usage of FL for a multi-robot exploration application. Instead of privacy, bandwidth and communication latency between the robots and the coordinating server are the dominant constraints. The robots are priceless assets (e.g., Mars rovers), and the operations team must do everything possible to avoid catastrophic accidents that would end the robots' mission. As the robots collectively learn models in their environment, mixed learning can be used to augment (from long distance) the robots' locally acquired data with helpful information. Given the limited transmit capacity, the desire is to send only that information which best complements the data the robots already possess. The targeted selection, collection, or simulation of useful `expert' centralized examples would be of utmost importance. 

Furthermore, as the robots' onboard datasets would be continually growing and changing in the course of their explorations of new environments, the nature of the centralized information transmitted to the robots would also need to change over time. In this continual learning scenario, the mixing of these two training data distributions would call for \emph{active} `expert' centralized data collection. 

In summary, the ability to measure and choose which examples of centralized data are most useful to aid a decentralized distribution in matching an inference distribution, and the ability to do this in a time-varying, `online' manner, are key next steps to unlocking further privacy and utility wins via FL.

%%%%%%%%%%%%%%%%%%%%%%%%%%%%%%%%%%%%%%%%%%%%%%%%%%%%%%%%%%%%

\bibliographystyle{plainnat}
\bibliography{fl_mixing}

%%%%%%%%%%%%%%%%%%%%%%%%%%%%%%%%%%%%%%%%%%%%%%%%%%%%%%%%%%%%

\newpage

\appendix

\section{Algorithms}
\label{app:algos}

Algorithms are provided for the mixing strategies described in Section~\ref{sec:strategies}. We first give the general $\fedavg$ algorithm (Algorithm~\ref{algo:generalized_fedavg}). We then show how it is modified to mix in a centralized data distribution, via Synchronous Parallel Training (Algorithm~\ref{algo:parallel_training}), Example Transfer (Algorithm~\ref{algo:ex_transfer}), or Gradient Transfer (Algorithm~\ref{algo:grad_transfer}). The highlighted \textcolor{blue}{blue text} indicates the steps added to basic $\fedavg$ in each case.

\begin{algorithm}[ht]
    \DontPrintSemicolon
    \SetKwInput{Input}{Input}
    \SetAlgoLined
    \LinesNumbered
    \begin{small}
    \Input{Initial model $\vx^{(0)}$; \clientopt, \serveropt with learning rate $\lr, \slr$}
    \For{$t \in \{0,1,\dots,T-1\}$ }{
      Sample a subset $\activeClients^{(t)}$ of clients\;
      \For{{\it \bf client} $i \in \activeClients^{(t)}$ {\it \bf in parallel}}{
        Initialize local model $\vx_i^{(t,0)}=\vx^{(t)}$\;
        \For {$k =0,\dots,\localStep_i-1$}{
          Sample a batch $\activeBatch_{i}^{(k)}$ of client data\;
          Compute local stochastic gradient w.r.t.~client batch $\sgrad_i(\vx_i^{(t,k)}; \activeBatch_{i}^{(k)})$\;
          Perform local update $\vx_i^{(t,k+1)} = \clientopt(\vx_i^{(t,k)}, \sgrad_i(\vx_i^{(t,k)}; \activeBatch_{i}^{(k)}), \lr, t)$\;
        }
        Compute local model changes $\localChange_i^{(t)} = \vx_i^{(t,\localStep_i)} - \vx_i^{(t,0)}$\;
      }
      Aggregate local changes $\localChange^{(t)} = \sum_{i \in \activeClients^{(t)}} p_i \localChange_i^{(t)} / \sum_{i \in \activeClients^{(t)}} p_i$\;
      Update global model $\vx^{(t+1)} = \serveropt(\vx^{(t)}, -\localChange^{(t)},\slr,t)$\;
    }
    \end{small}
    \caption{Generalized \fedavg~\citep{mcmahan2017fedavg, wang2021fedopt}}
    \label{algo:generalized_fedavg}
\end{algorithm}

\begin{algorithm}[ht]
    \DontPrintSemicolon
    \SetKwInput{Input}{Input}
    \SetAlgoLined
    \LinesNumbered
    \begin{small}
    \Input{Initial model $\vx^{(0)}$; \clientopt, \serveropt, \textcolor{blue}{\centralopt, \mergeopt}   with learning rate $\lr, \slr, \textcolor{blue}{\clr, \mlr}$}
    \For{$t \in \{0,1,\dots,T-1\}$ }{
      \textcolor{blue}{Initialize central model $x_c^{(t, 0)} = x^{(t)}$}\;
      \textcolor{blue}{\For{{\it \bf step} $j = 0,\dots,J$}{
          Sample a batch $\activeBatch^{(j)}$ of centralized data\;
          Compute stochastic gradient w.r.t.~centralized batch $\sgrad(\vx_c^{(t,j)}; \activeBatch^{(j)})$\;
          Perform update $\vx_c^{(t,j+1)} = \centralopt(\vx_c^{(t,j)}, \sgrad(\vx_c^{(t,j)}; \activeBatch^{(j)}), \clr, t)$\;
      }}
      \textcolor{blue}{Compute central model delta $\mergedChange_c^{(t)} = \vx_c^{(t,J)} - \vx^{(t)}$}\;
      Sample a subset $\activeClients^{(t)}$ of clients\;
      \For{{\it \bf client} $i \in \activeClients^{(t)}$ {\it \bf in parallel}}{
        Initialize local model $\vx_i^{(t,0)}=\vx^{(t)}$\;
        \For {$k =0,\dots,\localStep_i-1$}{
            Sample a batch $\activeBatch_{i}^{(k)}$ of client data\;
            Compute local stochastic gradient w.r.t.~client batch $\sgrad_i(\vx_i^{(t,k)}; \activeBatch_{i}^{(k)})$\;
            Perform local update $\vx_i^{(t,k+1)} = \clientopt(\vx_i^{(t,k)}, \sgrad_i(\vx_i^{(t,k)}; \activeBatch_{i}^{(k)}), \lr, t)$\;
        }
        Compute local model changes $\localChange_i^{(t)} = \vx_i^{(t,\localStep_i)} - \vx_i^{(t,0)}$\;
      }
      Aggregate local changes $\localChange^{(t)} = \sum_{i \in \activeClients^{(t)}} p_i \localChange_i^{(t)} / \sum_{i \in \activeClients^{(t)}} p_i$\;
      \textcolor{blue}{Compute federated model} $\textcolor{blue}{\vx_f^{(t)}} = \serveropt(\vx^{(t)}, -\localChange^{(t)},\slr,t)$\;
      \textcolor{blue}{Compute federated model delta $\mergedChange_f^{(t)} = \vx_f^{(t)} - \vx^{(t)}$}\;
      \textcolor{blue}{Aggregate central model and federated model deltas $\mergedChange^{(t)} = \alpha \mergedChange_c^{(t)} + (1 - \alpha) \mergedChange_f^{(t)}$}\;
      Update global model $\vx^{(t+1)} \mathbin{\textcolor{blue}{=}} \textcolor{blue}{\mergeopt(\vx^{(t)}, -\mergedChange^{(t)}, \mlr)}$\;
    }
    \end{small}     
    \caption{Generalized \fedavg \textcolor{blue}{in Parallel with Central Training}}
    \label{algo:parallel_training}
\end{algorithm}

\begin{algorithm}[ht]
    \DontPrintSemicolon
    \SetKwInput{Input}{Input}
    \SetAlgoLined
    \LinesNumbered
    \Input{Initial model $\vx^{(0)}$; \clientopt, \serveropt with learning rate $\lr, \slr$}
     \For{$t \in \{0,1,\dots,T-1\}$ }{
      Sample a subset $\activeClients^{(t)}$ of clients\;
      \For{{\it \bf client} $i \in \activeClients^{(t)}$ {\it \bf in parallel}}{
        \textcolor{blue}{Sample a batch $\activeBatch^{(t)}$ of centralized data, send to client $i$}\;
        \textcolor{blue}{Merge examples in $\activeBatch^{(t)}$ into existing client $i$ data}\;
        Initialize local model $\vx_i^{(t,0)}=\vx^{(t)}$\;
        \For {$k =0,\dots,\localStep_i-1$}{
            Sample a batch $\textcolor{blue}{\tilde{\activeBatch_{i}}}^{(k)}$ of \textcolor{blue}{mixed} data\;
            Compute local stochastic gradient w.r.t.~\textcolor{blue}{mixed} batch $\sgrad_i(\vx_i^{(t,k)}; \textcolor{blue}{\tilde{\activeBatch_{i}}}^{(k)})$\;
            Perform local update $\vx_i^{(t,k+1)} = \clientopt(\vx_i^{(t,k)}, \sgrad_i(\vx_i^{(t,k)}; \textcolor{blue}{\tilde{\activeBatch_{i}}}^{(k)}), \lr, t)$\;
        }
        Compute local model changes $\localChange_i^{(t)} = \vx_i^{(t,\localStep_i)} - \vx_i^{(t,0)}$\;
      }
      Aggregate local changes $\localChange^{(t)} = \sum_{i \in \activeClients^{(t)}} p_i \localChange_i^{(t)} / \sum_{i \in \activeClients^{(t)}} p_i$\;
      Update global model $\vx^{(t+1)} = \serveropt(\vx^{(t)}, -\localChange^{(t)},\slr,t)$\;
     }
     \caption{Generalized \fedavg \textcolor{blue}{with Example Transfer}}
     \label{algo:ex_transfer}
\end{algorithm}

\begin{algorithm}[ht]
    \DontPrintSemicolon
    \SetKwInput{Input}{Input}
    \SetAlgoLined
    \LinesNumbered
    \Input{Initial model $\vx^{(0)}$; \clientopt, \serveropt with learning rate $\lr, \slr$}
     \For{$t \in \{0,1,\dots,T-1\}$ }{
      \textcolor{blue}{Sample a batch $\activeBatch^{(t)}$ of centralized data}\;
      \textcolor{blue}{Compute stochastic gradient w.r.t.~centralized batch $\sgrad(\vx^{(t)}; \activeBatch^{(t)})$}\;
      Sample a subset $\activeClients^{(t)}$ of clients\;
      \For{{\it \bf client} $i \in \activeClients^{(t)}$ {\it \bf in parallel}}{
        Initialize local model $\vx_i^{(t,0)}=\vx^{(t)}$\;
        \textcolor{blue}{Set augmenting gradient $\tilde{\sgrad}^{(t)}=\sgrad(\vx^{(t)}; \activeBatch^{(t)})$}\;
        \For {$k =0,\dots,\localStep_i-1$}{
            Sample a batch $\activeBatch_{i}^{(k)}$ of client data\;
            Compute local stochastic gradient w.r.t.~client batch $\sgrad_i(\vx_i^{(t,k)}; \activeBatch_{i}^{(k)})$\;
            Perform local update $\vx_i^{(t,k+1)} = \clientopt(\vx_i^{(t,k)}, \sgrad_i(\vx_i^{(t,k)}; \activeBatch_{i}^{(k)}) \mathbin{\textcolor{blue}{+}} \textcolor{blue}{\tilde{\sgrad}^{(t)}}, \lr, t)$\;
        }
        Compute local model changes $\localChange_i^{(t)} = \vx_i^{(t,\localStep_i)} - \vx_i^{(t,0)}$\;
      }
      Aggregate local changes $\localChange^{(t)} = \sum_{i \in \activeClients^{(t)}} p_i \localChange_i^{(t)} / \sum_{i \in \activeClients^{(t)}} p_i$\;
      Update global model $\vx^{(t+1)} = \serveropt(\vx^{(t)}, -\localChange^{(t)},\slr,t)$\;
     }
     \caption{Generalized \fedavg \textcolor{blue}{with Gradient Transfer}}
     \label{algo:grad_transfer}
\end{algorithm}

\end{document}